\documentclass[letterpaper, 10pt, journal]{IEEEtran} 

\IEEEoverridecommandlockouts                              

\usepackage{graphicx} 
\usepackage{amsmath} 
\usepackage{amssymb}  
\usepackage{cite}
\usepackage{url}

\usepackage{algorithm}
\usepackage{algpseudocode}

\usepackage{booktabs}
\usepackage{amsthm}

\usepackage{xcolor}

\newcommand{\erev}{\color{purple}}

\theoremstyle{plain}

\newtheorem{proposition}{Proposition}

\theoremstyle{definition}

\usepackage{multirow}
\theoremstyle{remark}
\newtheorem{remark}{Remark}

\title{\LARGE \bf ProxPI: Proximal Prior Injection for Sampling-Based MPC \\ under Learned-Prior Mismatch}

\author{Euncheol Im$^{1,2}$, Myotaeg Lim$^{2}$ and Yisoo Lee$^{1,\dagger}$
\thanks{This work was supported by the National Research Foundation of Korea (NRF) grant funded by the Korea government(MSIT) (RS-2024-00339632) and by Hyundai Motor Company and Kia.}
\thanks{$^{1}$E. Im and Y. Lee are with the Center for Humanoid Research, Korea Institute of Science and Technology (KIST), 02792 Seoul, South Korea}
\thanks{$^{2}$E. Im and M. T. Lim are with the School of Electrical Engineering, Korea University, 02841 Seoul, South Korea}
\thanks{Corresponding author: Yisoo Lee {\tt\small yisoo.lee@kist.re.kr}.}
}

\begin{document}

\maketitle
\thispagestyle{empty}
\pagestyle{empty}

\begin{abstract}

Combining learned policies with model predictive control can leverage learned task priors while retaining online adaptation to new objectives and constraints, but performance degrades when the policy is out of distribution.
In policy-guided model predictive path integral (MPPI) control, a policy-centered warm-start approach centers the sampling distribution on the policy output. 
When the prior is mismatched, centering the sampling distribution on the policy output restricts exploration around an unsuitable solution and prevents recovery toward the task optimum.
We propose \emph{Proximal Prior Injection} (\emph{ProxPI}), which retains nominal-centered MPPI sampling and incorporates the policy through a soft proximity cost.
This matches the in-distribution performance of existing prior-injection schemes while enabling the optimizer to escape an inaccurate policy and recover vanilla MPPI-level performance.
We theoretically show that re-centering on the prior discards the optimizer's correction at every update, whereas nominal-centered sampling retains it and converges to a solution set by both the task cost and the prior, and that this failure is not removed by a larger rollout budget.
Simulations and real-robot experiments demonstrate robust performance under both in-distribution and out-of-distribution tasks.

\end{abstract}


\section{INTRODUCTION}

Learned policies and sampling-based model predictive control (MPC) offer complementary capabilities for real-time robot control, motivating efforts to integrate them within a unified framework. A learned policy, trained through reinforcement or imitation learning or represented by a generative model, can efficiently produce task-relevant actions using structure acquired during training. Its behavior, however, is shaped by its training objective and distribution, and the policy alone does not explicitly re-optimize newly specified objectives or constraints at deployment. Sampling-based MPC, particularly model predictive path integral (MPPI) control~\cite{williams2017model,williams2018information}, instead optimizes deployment-time trajectory costs through model rollouts without requiring analytic derivatives of the dynamics or cost, but its finite-sample performance depends critically on how candidate control sequences are proposed.

\begin{figure}[t]
    \centering
    \includegraphics[width=0.95\columnwidth]{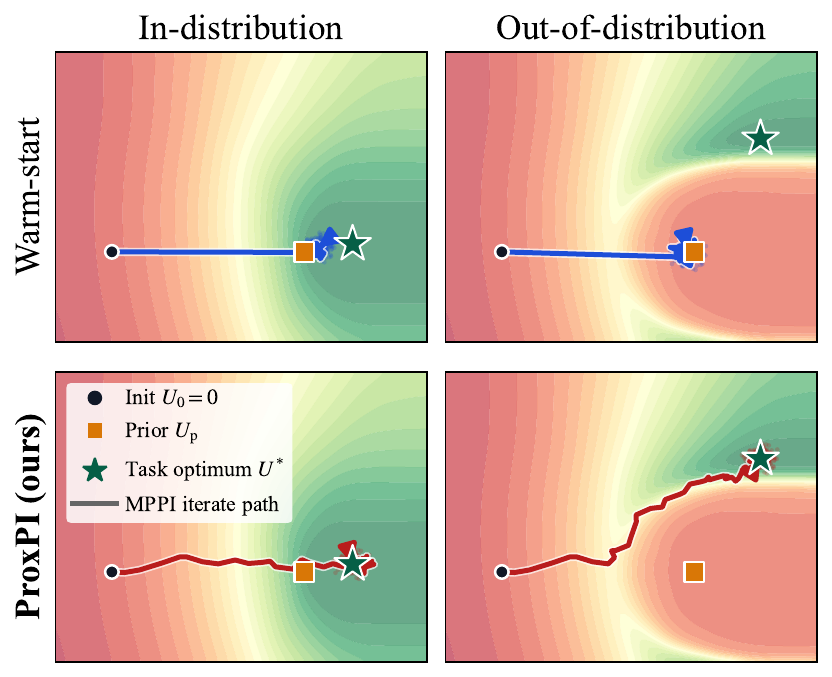}
    \caption{Conceptual illustration of prior injection in control-sequence space.
    The background is the cost landscape $J(U)$ (green low, red high), and each panel overlays the MPPI iterate path from $U_0{=}0$ and the resulting sample cloud.
    Rows are warm-start and ProxPI (ours), columns are in- and out-of-distribution.
    In distribution the prior $U_{\mathrm{p}}$ coincides with the task optimum $U^{*}$ and both schemes converge there, whereas out of distribution the optimum moves to $U^{*}$ while $U_{\mathrm{p}}$ stays in a now high-cost region, so warm-start is pinned to $U_{\mathrm{p}}$ while ProxPI continues optimizing toward $U^{*}$.} 
    \label{fig:overview}
\end{figure}

These approaches can be coupled at different stages of the control pipeline. MPC can generate supervision for policy learning~\cite{zhang2016mpcgps}, while a learned policy can be applied as a feedback residual to a model-based command~\cite{johannink2019residual,cheng2025rambo,jeon2025residual}. A growing line of work instead incorporates a learned policy or generative model into the online optimizer as a learned proposal, or sampling prior, allowing learned structure to guide the search while sampling-based MPC evaluates deployment-specific objectives and cost-encoded constraints. Such proposals have been represented by learned policies, including importance-sampling policies~\cite{carius2022constrained} and reinforcement learning (RL) policies~\cite{qu2024rldriven,kotecha2025real,seo2026rgb,liang2026policy}, as well as by normalizing flows~\cite{power2024learning,sacks2023learning}, diffusion models~\cite{huang2024diffusion}, and flow-matching models~\cite{kurtz2025generative,brudermuller2026generative}. Existing schemes learn and update a structured proposal directly~\cite{power2024learning,sacks2023learning}, augment samples from the online optimizer with learned proposals~\cite{trevisan2024biased}, or initialize the proposal from a policy-generated sequence~\cite{qu2024rldriven,seo2026rgb}. When the learned proposal is aligned with the deployment task, these mechanisms concentrate computation on promising regions and can substantially reduce the rollout budget required for a high-quality solution.

This advantage, however, depends on the learned proposal remaining aligned with the deployment task. Even policies trained with broad randomization or adaptation are optimized over prescribed families of environments, commands, dynamics, and behaviors~\cite{hwangbo2019learning,kumar2021rma,rudin2022learning}; an objective outside that family can make the proposed control sequence 
far from the task optimum. For example, a velocity-tracking locomotion policy may provide little useful guidance for a qualitatively different behavior such as a sustained squat or bipedal rearing. When the learned component shapes the rollout proposal, such mismatch allocates a finite sampling budget toward task-irrelevant regions. Prior work mitigates distribution shift by projecting an out-of-distribution (OOD) environment representation toward the training distribution~\cite{power2024learning} or by combining learned proposals with samples adapted by the online optimizer~\cite{brudermuller2026generative}. Related policy-customization methods retain prior behavior through policy-likelihood or action-deviation terms while also using the policy to initialize or guide sampling~\cite{wang2025residualmppi,kotecha2025real}. Nevertheless, the effect of repeatedly replacing MPPI's nominal sampling center with a mismatched policy output remains insufficiently characterized.

We focus on this strongest form of coupling, which we call \emph{policy-centered warm-start}: at every MPC update, the sampling mean is replaced by the policy-generated control sequence. Under prior mismatch, the rollout distribution is repeatedly concentrated around a region that may lie far from the task optimum, leaving task-relevant regions with negligible coverage under a finite rollout budget. Moreover, the corrected nominal obtained by MPPI is overwritten at the next update, preventing corrective shifts in the proposal mean from accumulating across control steps. The resulting failure is therefore structural to persistent re-centering rather than merely a shortage of local exploration. Increasing the sampling variance $\sigma$ broadens the search around the same mismatched center and, within practical finite-sample regimes, does not reliably restore coverage of the task optimum.

To address this limitation, we propose \emph{Proximal Prior Injection} (\emph{ProxPI}), which keeps MPPI sampling centered at its evolving nominal sequence and incorporates the learned prior only through a proximal term in the trajectory cost. The prior therefore acts as a soft, escapable preference rather than a fixed sampling center. With a moderate proximal weight, an aligned prior biases the optimizer toward promising control sequences; under mismatch, the task objective retains a mechanism to move the sampling center away from the prior. Setting the proximal weight to zero exactly recovers vanilla MPPI. Under a local Gaussian approximation, we show that the sampling-center contribution to the $\chi^2$ divergence of policy-centered warm-start grows exponentially with prior displacement, whereas the corresponding contribution for the proposed method is governed by the displacement of the carried MPPI nominal. We separately characterize how the proximal weight shifts and contracts the effective target distribution.

In summary, this paper makes the following contributions:
\begin{itemize}
    \item \textbf{Proximal Prior Injection}: 
    We propose a policy-guided MPPI formulation that decouples prior guidance from the sampling center, keeping the sampling distribution centered at the evolving MPPI nominal while the learned prior enters only through a proximal cost, allowing the optimizer to exploit the prior without being confined to it.

    \item \textbf{Theoretical robustness analysis}: 
    We characterize how the prior-injection site affects repeated MPPI updates under a local Gaussian approximation. Nominal-centered updates retain corrections across iterations, whereas policy-centered warm-start repeatedly re-centers the proposal and discards them.
    We further show that finite-sample weight concentration is a separate mechanism, so increasing the rollout budget does not remove the re-centering failure.
    
    \item \textbf{Multi-platform validation}: 
    Simulations and experiments on a Summit XLS holonomic mobile robot, a planar 3-DoF arm, an FR3 manipulator, a Go2 quadruped, and a G1 humanoid show that the proposed method matches the other injection schemes under an aligned prior while remaining robust to prior mismatch, and a comparison against those schemes shows that the failure is consistently associated with displacing every rollout away from the nominal rather than with any particular scheme.
\end{itemize}

\section{Preliminaries: Model Predictive Path Integral}
\label{sec:prelim}

MPPI~\cite{williams2017model,williams2018information} is a sampling-based stochastic optimal control method that avoids differentiating the dynamics and cost.
It evaluates many randomly perturbed control sequences in parallel and fuses them through an importance-weighted average.
We review MPPI from an information-theoretic viewpoint, which makes explicit (i) the optimal \emph{distribution} that the algorithm estimates and (ii) the quantity that governs the quality of that estimate.

\subsection{Stochastic Optimal Control}
We consider a system whose discrete-time dynamics are
\begin{equation}
    \mathbf{x}_{t+1} = f(\mathbf{x}_t, \mathbf{v}_t),
\label{eq:dyn}
\end{equation}
where $\mathbf{x}_t \in \mathbb{R}^{n_x}$ is the state and
$\mathbf{v}_t \in \mathbb{R}^{n_u}$ the applied control input.
Over a horizon of $T$ steps, the decision variable is the nominal control sequence
\begin{equation}
    U = (\mathbf{u}_0, \mathbf{u}_1, \dots, \mathbf{u}_{T-1}),
\label{eq:nominal}
\end{equation}
which MPPI maintains as its current estimate of the optimal control and uses as the center around which candidate inputs are sampled.
Each applied control is the nominal input perturbed by zero-mean Gaussian noise,
\begin{equation}
    \mathbf{v}_t = \mathbf{u}_t + \delta\mathbf{u}_t, \qquad
    \delta\mathbf{u}_t \sim \mathcal{N}(\mathbf{0}, \Sigma),
\label{eq:mppi_input}
\end{equation}
so that perturbing $U$ yields a rollout $V = (\mathbf{v}_0, \dots, \mathbf{v}_{T-1})$.
Each rollout is scored by the trajectory cost
\begin{equation}
    S(V) = \phi(\mathbf{x}_T) + \sum_{t=0}^{T-1}\ell(\mathbf{x}_t, \mathbf{v}_t),
\label{eq:mppi_cost}
\end{equation}
with terminal cost $\phi$ and running cost $\ell$ (e.g., tracking error, control effort, or constraint penalties). 
The stochastic optimal control problem seeks the nominal sequence that minimizes the expected cost,
\begin{equation}
    \min_{U}\ \mathbb{E}_{V}\!\left[S(V)\right].
\label{eq:socp}
\end{equation}

\subsection{Information-Theoretic Solution}

Following~\cite{williams2018information} and the probabilistic-inference formulation summarized in~\cite{honda2026inference}, we relax \eqref{eq:socp} to the free-energy problem $\min_{p}\ \mathbb{E}_{p}[S(V)]+\lambda\,D_{\mathrm{KL}}(p\,\Vert\,p_0)$, whose minimizer is the Gibbs distribution
\begin{equation}
    p^*(V) \;\propto\; \exp\!\big(-S(V)/\lambda\big)\, p_0(V),
\label{eq:mppi_optdist}
\end{equation}
where $\lambda>0$ is the temperature and $p_0$ is the base distribution induced by the nominal sequence through \eqref{eq:mppi_input}, i.e., $V\sim\mathcal{N}(U,\Sigma)$.  This base distribution is exactly the Gaussian distribution $q$ from which rollouts are sampled, so $q(V)=p_0(V)$.

The MPPI update targets the mean of this distribution, $\mathbb{E}_{p^*}[V]$.
This expectation is generally intractable for the nonconvex trajectory costs considered here, so we estimate it from
$K$ rollouts $\{V_k\}_{k=1}^K$ drawn from $q$, approximating
the mean by the importance-weighted average
\begin{subequations}\label{eq:mppi_update}
\begin{align}
    \omega_k &= \exp\!\big(-\tilde{S}_k/\lambda\big), \label{eq:mppi_update_w}\\
    \mathbf{u}^{*}_t &\leftarrow \mathbf{u}_t
      + \frac{\sum_{k=1}^K \omega_k\,\delta\mathbf{u}_{k,t}}
             {\sum_{k=1}^K \omega_k}, \label{eq:mppi_update_u}
\end{align}
\end{subequations}
where $\tilde{S}_k = S_k - \min_{j\in\{1,\dots,K\}} S_j$ shifts the costs across the $K$ rollouts for numerical stability without altering the weight ratios.
Equation~\eqref{eq:mppi_update} is precisely the self-normalized importance-sampling estimate of $\mathbb{E}_{p^*}[V]$. 
The normalized weight $\omega_k/\sum_{j=1}^{K}\omega_j$ equals $w(V_k)/\sum_{j=1}^{K} w(V_j)$ with $w=p^*(V)/q(V)$, and because the choice $q(V)=p_0(V)$ cancels the base-measure factor in~\eqref{eq:mppi_optdist}, the weight reduces to a function of the trajectory cost alone, $w(V)\propto\exp(-S(V)/\lambda)$, recovering the cost-based weights in~\eqref{eq:mppi_update_w}.

\subsection{Quality of the Sample-Based Estimate}
\label{sec:estimate_quality}
Because \eqref{eq:mppi_update} is an importance-sampling estimate, its reliability under a finite rollout budget $K$ depends on how well the sampling distribution $q$ covers the optimal distribution $p^*$.
When a few rollouts carry most of the weight, the number of effectively contributing rollouts collapses and the estimate becomes noisy.
This concentration is captured exactly by the chi-squared ($\chi^2$) divergence between target and sampling distribution~\cite{kong1992note, agapiou2017importance},
\begin{equation}
    \frac{\mathrm{Var}_q[w]}{(\mathbb{E}_q[w])^2}
    = \chi^2\!\big(p^*\,\Vert\,q\big), \qquad w=\frac{p^*(V)}{q(V)},
\label{eq:mppi_chi2}
\end{equation}
which we adopt in Sec.~{\ref{sec:finite_sample} as the measure of importance-weight degeneracy, where smaller $\chi^2$ means more rollouts contribute and $\chi^2=0$ if and only if $q(V)=p^*(V)$.
At the distribution level, this degeneracy corresponds to an effective sample
size~(ESS)~\cite{kong1992note} of $K/(1+\chi^{2})$, which equals $K$ when $q=p^*$.
From a finite rollout set, we compute the corresponding empirical ESS directly
from the weights $\omega_k$,
\begin{equation}
    \widehat{\mathrm{ESS}} \;=\;
    \frac{\big(\sum_{k}\omega_k\big)^2}{\sum_{k}\omega_k^{2}}.
\label{eq:ess}
\end{equation}
It can be interpreted as the number of rollouts contributing as if equally
weighted, and lies in $[1,K]$, attaining $K$ when the weights are equal and $1$
when a single rollout carries all the weight.

\section{Proposed Method}
This section analyzes how the site of prior injection affects the evolution of
sampling-based MPC updates. Under a local Gaussian approximation, nominal-centered
updates retain and accumulate task-driven corrections toward a fixed point determined
jointly by the task cost and the prior, whereas policy-centered warm-start repeatedly
re-centers the proposal and therefore does not carry those corrections across updates.
We then show that finite-sample weight concentration constitutes a separate mechanism
induced by proposal--target displacement, and characterize how the proximal weight
controls the convergence--bias trade-off.

\subsection{Warm-Start and Proximal Prior Injection}
A learned prior provides a candidate control sequence $U_{\mathrm{p}}$ for the current state.
We consider two ways of injecting $U_{\mathrm{p}}$ into the MPPI update of Sec.~\ref{sec:prelim}.
The first, used by previous work, relocates the sampling distribution so that rollouts are perturbations of the prior rather than of the nominal sequence,
\begin{equation}
    q_{\mathrm{w}}(V) = \mathcal{N}(U_{\mathrm{p}}, \Sigma).
\label{eq:warmstart}
\end{equation}

The second, which we adopt, leaves the baseline sampling distribution $q=\mathcal{N}(U,\Sigma)$ unchanged and instead augments the cost with a quadratic residual that penalizes deviation from the prior,
\begin{equation}
    S'(V) = S(V) + \alpha\,\lVert V - U_{\mathrm{p}}\rVert^2,
\label{eq:proxpi}
\end{equation}
where $\alpha>0$ weights the prior.
We refer to \eqref{eq:warmstart} as \emph{warm-start} and to \eqref{eq:proxpi} as \emph{ProxPI}.

\begin{remark}[Connection to warm-start]
\label{rem:duality}
Substituting $S'$ into \eqref{eq:mppi_optdist}, the ProxPI target factorizes as
\begin{equation}
    p'(V) \propto \exp\!\big(-S(V)/\lambda\big)\,
    \exp\!\big(-\tfrac{\alpha}{\lambda}\lVert V-U_{\mathrm{p}}\rVert^2\big)\,p_0(V),
\end{equation}
the product of the MPPI likelihood and a Gaussian prior centered at $U_{\mathrm{p}}$ with covariance $\tfrac{\lambda}{2\alpha}I$.
ProxPI and warm-start therefore both inject a Gaussian prior centered at $U_{\mathrm{p}}$, one through the target~\eqref{eq:proxpi} and the other through the sampling distribution~\eqref{eq:warmstart}.
The two share this center but not their covariance.
Warm-start's is the sampling covariance $\Sigma$, whereas ProxPI's is $\tfrac{\lambda}{2\alpha}I$, set independently by the weight $\alpha$ and coinciding with $\Sigma$ only when $\tfrac{\lambda}{2\alpha}I=\Sigma$.
The analysis below characterizes how this difference in injection site and width, rather than the prior center, affects repeated nominal updates and finite-sample weight concentration.
\end{remark}

\subsection{Local Nominal Dynamics}
\label{sec:local_dynamics}

We first isolate how the prior-injection site changes the evolution of the
MPPI nominal. Around a local minimizer $U^*$ of the task cost $S$, we expand the
ProxPI objective~\eqref{eq:proxpi}, approximating its task part alone, to obtain
the local model
\begin{equation}
S'(V)
\simeq
S(U^*)
+\frac{1}{2}\|V-U^*\|_H^2
+\alpha\|V-U_{\mathrm p}\|^2,
\label{eq:local_cost}
\end{equation}
where $\|x\|_H^2\triangleq x^\top Hx$ and $H\succeq0$. Setting $\alpha=0$ leaves the
task cost that vanilla MPPI and warm-start optimize.
We treat $U^*$, $H$, $U_{\mathrm p}$, and $\Sigma$ as fixed across updates,
removing their state dependence so that the repeated-update mechanism can be
examined on its own.

To cover both schemes with one expression, write the proposal center as $\bar U$,
so that $q(V)=\mathcal{N}(\bar U,\Sigma)$. Substituting \eqref{eq:local_cost} into
the Gibbs target in~\eqref{eq:mppi_optdist} gives
\begin{equation}
p_\alpha(V)=\mathcal{N}\!\left(m_\alpha(\bar U),C_\alpha\right),
\label{eq:local_target}
\end{equation}
where
\begin{align}
C_\alpha
&=
\left[
\Sigma^{-1}+\frac{H+2\alpha I}{\lambda}
\right]^{-1},
\label{eq:local_cov}\\
m_\alpha(\bar U)
&=
C_\alpha
\left[
\Sigma^{-1}\bar U
+\frac{1}{\lambda}HU^*
+\frac{2\alpha}{\lambda}U_{\mathrm p}
\right].
\label{eq:local_mean}
\end{align}
Here, $m_\alpha(\bar U)$ is the infinite-sample limit of the
MPPI estimate in~\eqref{eq:mppi_update}, with fixed point
\begin{equation}
U^\dagger
=
(H+2\alpha I)^{-1}
(HU^*+2\alpha U_{\mathrm p}).
\label{eq:prox_fixed_point}
\end{equation}
Whether the updates approach $U^\dagger$ depends on the center supplied at each
step. Nominal-centered sampling supplies the previous nominal, $\bar U=U_j$,
whereas policy-centered warm-start supplies $\bar U=U_{\mathrm p}$ at every update.
The derivation of \eqref{eq:local_cov}--\eqref{eq:prox_fixed_point} is given in
Appendix~\ref{app:local_target}.

\begin{proposition}[Local nominal dynamics]
\label{prop:contraction}
Let $H+2\alpha I\succ0$ and set $A_\alpha=C_\alpha\Sigma^{-1}$. Then
$\rho(A_\alpha)<1$, and under nominal-centered sampling, $U_{j+1}=m_\alpha(U_j)$
satisfies
\begin{equation}
U_{j+1}-U^\dagger=A_\alpha(U_j-U^\dagger),
\label{eq:prox_contraction}
\end{equation}
so $U_j\rightarrow U^\dagger$ geometrically from any initial nominal $U_0$. Under
policy-centered warm-start, which carries no proximal term, the update returns
$m_0(U_{\mathrm p})$ at every $j$, independently of the updates preceding it.
\end{proposition}
The contraction follows because $C_\alpha\prec\Sigma$, as \eqref{eq:local_cov} shows
directly; the proof is given in Appendix~\ref{app:contraction}.
Related contraction results for standard MPPI under quadratic costs have been
established in~\cite{yi2024covo}.

\begin{remark}[What re-centering removes]
Re-centering does not eliminate the one-step MPPI correction. At $\alpha=0$, for any
$H\succeq0$,
\begin{equation}
m_0(U_{\mathrm p})-U^*
=
A_0(U_{\mathrm p}-U^*),
\qquad
A_0=C_0\Sigma^{-1}.
\label{eq:warm_reset}
\end{equation}
The eigenvalues of $A_0$ lie in $(0,1]$, and strictly below one along directions
in which $H$ has positive curvature, so each update still moves toward $U^*$ there.
What re-centering removes is the carrying of that correction into the next update.
\end{remark}

\subsection{Finite-Sample Weight Degeneracy}
\label{sec:finite_sample}

With finite rollout budgets, proposal--target displacement can additionally
degrade the reliability of the MPPI update by concentrating importance weight on a
small subset of samples. To characterize this effect separately from persistent
re-centering, write $\bar U_j$ for the proposal center at
update $j$, where $\bar U_j=U_j$ for ProxPI and vanilla MPPI, whereas
$\bar U_j=U_{\mathrm p}$ for policy-centered warm-start.
The resulting estimate is
$\widehat U_{j+1}=m_\alpha(\bar U_j)+\varepsilon_j$,
where $\varepsilon_j$ is the finite-$K$ importance-sampling error.

Following Sec.~\ref{sec:estimate_quality}, we use the $\chi^2$ divergence as a
distribution-level measure of the weight degeneracy behind $\varepsilon_j$.
It also appears in non-asymptotic importance-sampling
error bounds~\cite{agapiou2017importance}, though it is not the exact variance of
the self-normalized MPPI mean estimator. Because each scheme has its own local
Gibbs target, the divergence measures how reliably a scheme estimates its own
update, not whether that update is good for the task.

Suppressing the update index, for
$q(V)=\mathcal{N}(\bar U,\Sigma)$ and
$p_\alpha(V)=\mathcal{N}(m_\alpha(\bar U),C_\alpha)$,
\begin{equation}
\chi^2(p_\alpha\|q)+1
=
\frac{|\Sigma|}
{\sqrt{|C_\alpha|\,|2\Sigma-C_\alpha|}}
\exp\!\left[
g_\alpha^\top
(2\Sigma-C_\alpha)^{-1}
g_\alpha
\right],
\label{eq:general_chi2}
\end{equation}
where
\begin{equation}
g_\alpha
=
m_\alpha(\bar U)-\bar U
=
\frac{C_\alpha}{\lambda}
\left[
H(U^*-\bar U)
+
2\alpha(U_{\mathrm p}-\bar U)
\right].
\label{eq:g_alpha}
\end{equation}
The derivation is given in Appendix~\ref{app:gaussian_chi2}.
Since $H\succeq0$ and $\alpha\geq0$, \eqref{eq:local_cov} gives
$C_\alpha\preceq\Sigma$, so $2\Sigma-C_\alpha$ is positive definite and the
divergence is finite.

Equation~\eqref{eq:general_chi2} separates the proposal--target mean
displacement from the covariance mismatch, which Sec.~\ref{sec:alpha_tradeoff}
takes up. For ProxPI, $\bar U=U_j$, so $g_\alpha$ is the deterministic step
$m_\alpha(U_j)-U_j$ of Sec.~\ref{sec:local_dynamics}, which vanishes as $U_j$
approaches the fixed point $U^\dagger$. For policy-centered warm-start,
$\alpha=0$ and $\bar U=U_{\mathrm p}$, giving
$g_{\mathrm w}=(C_0/\lambda)H(U^*-U_{\mathrm p})$, which reappears at
every reset under the frozen local model. Through \eqref{eq:general_chi2}, the
mean-displacement contribution to the ProxPI divergence therefore vanishes as its
nominal converges, whereas the warm-start contribution stays at the level set by
$U^*-U_{\mathrm p}$.

Together with Sec.~\ref{sec:local_dynamics}, this identifies two distinct
mechanisms by which policy-centered warm-start fails under prior mismatch. The
first is weight concentration, where a large proposal--target displacement makes
the finite-$K$ correction unreliable. 
Related sampling-complexity bounds for finite-sample path-integral estimation
have been derived in~\cite{yoon2022sampling}. The second is persistent re-centering, which
discards the estimated correction before the next update. Being present already in
the infinite-sample limit, the second is not removed by increasing the rollout
budget.

\subsection{Proximal-Weight Trade-off}
\label{sec:alpha_tradeoff}

The proximal weight $\alpha$ controls a trade-off between local convergence
and prior bias. From~\eqref{eq:prox_fixed_point},
\begin{equation}
U^\dagger-U^*
=
(H+2\alpha I)^{-1}
2\alpha(U_{\mathrm p}-U^*),
\label{eq:prox_bias}
\end{equation}
so increasing $\alpha$ moves the fixed point toward a mismatched prior.

Equation~\eqref{eq:local_cov} simultaneously shows that increasing $\alpha$
contracts $C_\alpha$. Hence, even when $g_\alpha$ vanishes at
$U^\dagger$, the covariance-mismatch term in~\eqref{eq:general_chi2} can
still concentrate the importance weights.

For isotropic sampling $\Sigma=r^2I$, let $h_i$ be an eigenvalue of $H$ and
consider its eigendirection. Specializing $A_\alpha$ of
Proposition~\ref{prop:contraction} and the coefficient in~\eqref{eq:prox_bias} to
that direction gives
\begin{equation}
a_i
=
\frac{\lambda}
{\lambda+r^2(h_i+2\alpha)},
\qquad
b_i
=
\frac{2\alpha}{h_i+2\alpha}.
\label{eq:alpha_factors}
\end{equation}
Here $a_i$ is the fraction of the distance to $U^\dagger$ that survives one update,
and $b_i$ is the fraction of the prior displacement $U_{\mathrm p}-U^*$ that
remains at $U^\dagger$.

Increasing $\alpha$ therefore lowers $a_i$ and raises $b_i$, so faster local
contraction is paid for with bias toward the prior. Together with the weight
concentration above, this explains why overly strong proximal guidance can reduce
robustness under mismatch.

\section{Experiments}
\label{sec:exp}
We evaluate ProxPI and six prior-injection baselines, together with vanilla MPPI, on five robotic platforms in simulation, and validate the warm-start and ProxPI contrast on a real Franka FR3 arm.
Residual-MPPI is evaluated on the two low-dimensional tasks only, since it requires an action likelihood that the remaining priors do not provide.
Unless otherwise stated, all simulated results are based on \(N=50\) runs, and task-progress scores are reported as mean \(\pm\) standard deviation.

\subsection{Experimental Setup}
All controllers use MPPI with MuJoCo dynamics, with the three higher-DoF
platforms implemented in MuJoCo MPC (MJPC)~\cite{howell2022predictive}; their
joint-position targets are tracked by low-level PD controllers.
Table~\ref{tab:config} summarizes the task-specific MPPI parameters; additional
implementation details are provided in Appendix~\ref{app:exp_details}, and complete
task costs in the supplementary material.

\begin{table}[tbp]
\centering
\caption{MPPI Parameters per Simulated Task}
\label{tab:config}
\begin{tabular}{lcccccc}
\toprule
Task & {$n_u$} & $K$ & $T$ & $\sigma$ & $\lambda$ & $\alpha$ \\
\midrule
2D navigation  & {3}  & 64  & 20  & 0.15  & 0.1  & 2.0 \\
2D reaching    & {3}  & 64  & 20  & 0.10  & 0.1  & 0.1 \\
FR3 obstacle   & {7}  & 32  & 20  & {$0.04$--$0.05$\textsuperscript{$\ddagger$}}   & {30}  & 1 \\
Go2 rearing    & {12} & 128 & 40  & 0.10  & {10\textsuperscript{$\dagger$}}  & 5 \\
G1 squat       & {29} & 64  & 40  & {$0.01$--$0.03$\textsuperscript{$\ddagger$}}  & {0.5}  & {0.1} \\
\bottomrule
\end{tabular}
{\par\vspace{2pt}\footnotesize\raggedright
$\dagger$ Each scheme is run at its own best $\lambda$. This coincides for every
scheme on all tasks except the Go2, where ProxPI's optimum is $\lambda{=}2$ and
the remaining five schemes share $\lambda{=}10$.
$\ddagger$ A range denotes per-joint values.\par}
\end{table}

\textbf{Environment mismatch.}
We evaluate this setting on 2D navigation, 2D reaching, and the simulated FR3
by introducing obstacles absent from prior training. The two low-dimensional tasks use
obstacle-free SAC priors with planar commands $(v_x,v_y,\omega)$ and joint-torque
actions, respectively;
their in-distribution~(ID) evaluation retains the training environment, whereas the
OOD evaluation blocks the direct route to the target with an unseen
obstacle. The FR3 uses joint-position targets, a flow-matching prior trained on obstacle-free
inverse-kinematics reaching trajectories, and single-instance
sampling~\cite{kim2025single}, with the obstacle similarly obstructing a commanded
6-DoF end-effector reach.

\textbf{Behavior mismatch.}
We evaluate this setting on the Go2 and G1 using velocity-tracking
reinforcement-learning locomotion priors from Unitree RL Lab~\cite{unitree_rl_lab}.
With the environment unchanged, the Go2 is commanded to an increased base height and
nose-up pitch, requiring it to rear, whereas the G1 is commanded to a lowered pelvis
height with the torso upright, requiring a sustained squat; both motions lie outside
the respective prior objectives.

The prior sequence $U_{\mathrm p}$ follows each model's native output structure. The
SAC policies are rolled out in closed loop over the prediction horizon, the
flow-matching prior predicts the full joint-position sequence in one query, and the
current Go2 and G1 policy outputs are held constant over the horizon.

On the real FR3, an ACT behavioral-cloning prior~\cite{zhao2023learning} trained for a
single end-effector target predicts the full joint-position sequence without observing
the commanded target. After convergence to the training target, the command is
switched to an unseen location, creating target-induced prior mismatch. Training and
protocol details are given in Appendix~\ref{app:exp_details}.

\subsection{Prior-Injection Schemes}
Table~\ref{tab:schemes} summarizes the sampling-side and objective-side structures of
the compared schemes.
Mixture-Elite applies elite selection to the same mixed rollout pool, placing it
between Mixture and CEM-style elite updates. For GPC-CEM, the original method
bootstraps its generative prior from a large corpus of sampling-based MPC solutions;
here, prior samples are instead drawn from the same learned prior used by the other
schemes to isolate the effect of the injection mechanism. Residual-MPPI centers
sampling on $U_{\mathrm p}$ and replaces the original task cost with its residual
objective augmented by the prior action log-likelihood, so that the objective retains
only the newly added requirement, which on these tasks is the obstacle-avoidance term.

\begin{table}[tbp]
\centering
\caption{Prior-Injection Schemes}
\label{tab:schemes}
\small
\setlength{\tabcolsep}{3.5pt}
\begin{tabular}{@{}lll@{}}
\toprule
Scheme & Sampling-side injection & Objective \\
\midrule
Vanilla MPPI  & $U$-centered                                 & $S$ \\
Warm-start    & $U_{\mathrm p}$-centered                     & $S$ \\
\multirow{2}{*}{Mixture} & $K/2$ at $U$             & \multirow{2}{*}{$S$} \\
              & $K/2$ at $U_{\mathrm p}$                     &     \\
Mixture-Elite & Mixture $+$ elite                            & $S$ \\
Blend         & $\ell U+(1-\ell)U_{\mathrm p}$                & $S$ \\
GPC-CEM       & CEM $+$ prior samples                        & $S$\textsuperscript{$\dagger$} \\
Residual-MPPI & $U_{\mathrm p}$-centered                     & residual $+$ log-likelihood \\
ProxPI        & $U$-centered                                 & $S+\alpha\lVert V-U_{\mathrm p}\rVert^2$ \\
\bottomrule
\end{tabular}
\par\vspace{2pt}\footnotesize\raggedright
Warm-start follows~\cite{seo2026rgb};
Mixture~\cite{power2024learning,kurtz2025generative},
Blend~\cite{huang2024diffusion},
GPC-CEM~\cite{brudermuller2026generative},
Residual-MPPI~\cite{wang2025residualmppi}.
Blend uses $\ell{=}0.5$.
$\dagger$ GPC-CEM executes the lowest-cost candidate rather than a weighted mean.\par
\end{table}

\begin{figure*}[tbp]
\centering
\includegraphics[width=0.9\textwidth]{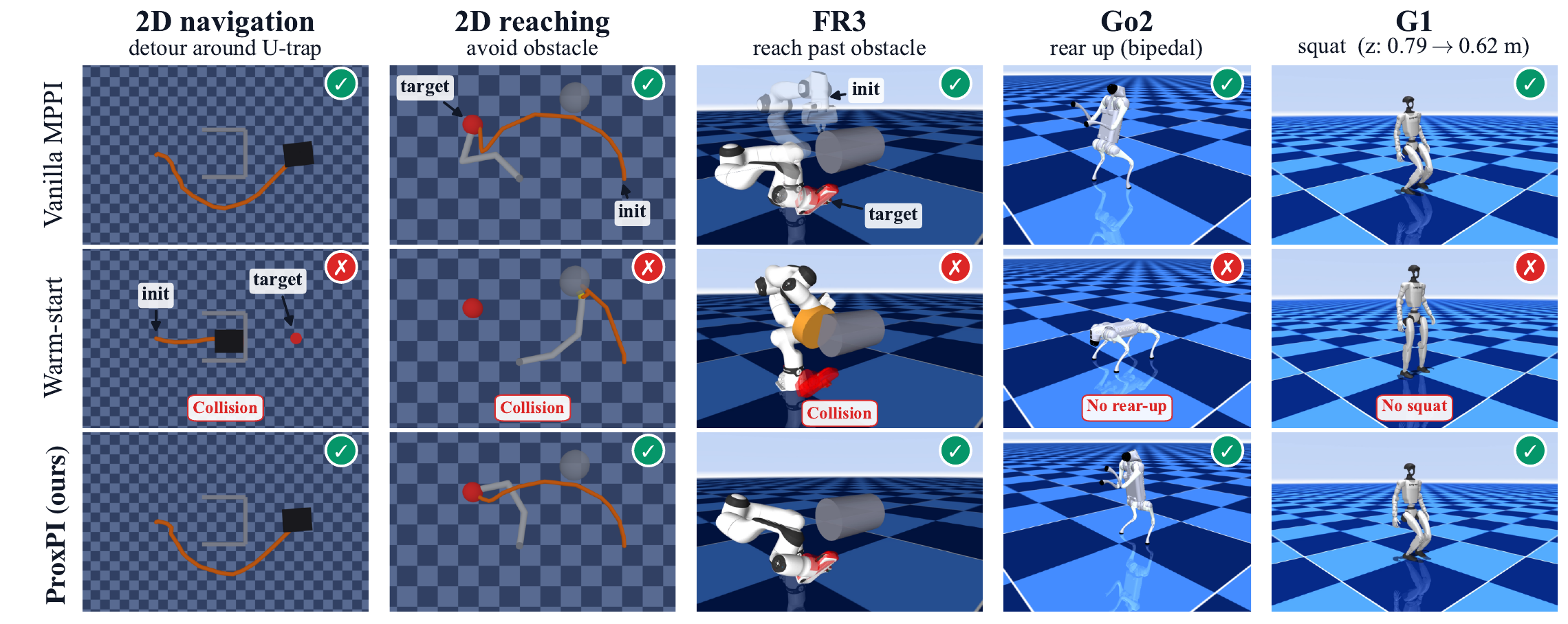}
\caption{
Task behaviors under OOD prior mismatch across five robots.
Columns denote tasks and rows denote vanilla MPPI, warm-start, and ProxPI, the prior-free reference and the two extremes of where the prior can enter, the sampling center and the cost.
The first two columns show executed trajectories (orange), whereas the remaining columns show terminal configurations.
Check and cross markers denote success and failure, with the failure reason shown in each unsuccessful panel.
Under prior mismatch, warm-start fails all tasks, whereas vanilla MPPI and ProxPI complete them.
}
\label{fig:ood}
\end{figure*}

\subsection{Task-Progress Metrics}
Although all platforms are controlled to reach and maintain their respective task targets, the targets are defined in different task spaces. We therefore design a platform-specific progress metric
$p\in[0,1]$ for the simulated evaluations, where $p=1$ denotes that the target is reached and maintained and $p=0$ denotes no progress or task failure.
Unless a collision occurs, all metric components are averaged over the final $25\%$ of the episode so that the score reflects sustained task completion rather than a transient configuration. The operator $\operatorname{clip}_{[0,1]}(\cdot)$ limits each metric to $[0,1]$.

\subsubsection{Collision-Free Goal Reaching}
For the mobile robot XLS and planar arm, let $e_{\mathrm p}$ denote the current task-space position error and $e_{\mathrm p}^{0}$ its initial value. Their progress metrics are
\begin{align}
p_{\mathrm{nav}}=p_{\mathrm{arm}}
=
\begin{cases}
0, & \text{if collision},\\[1mm]
\operatorname{clip}_{[0,1]}
\left(1-\dfrac{e_{\mathrm p}}{e_{\mathrm p}^{0}}\right),
& \text{otherwise}.
\end{cases}
\label{eq:progress_lowdof}
\end{align}
Thus, motion toward the target increases the score, while any obstacle contact is treated as task failure.

The simulated FR3 task additionally requires end-effector orientation tracking. Let $e_{\theta}$ and $e_{\theta}^{0}$ denote the current and initial orientation errors, respectively. The FR3 progress is defined as
\begin{align}
p_{\mathrm{FR3}}
=
\begin{cases}
0, & \text{if collision},\\[1mm]
\operatorname{clip}_{[0,1]}
\left[
\min\left(
1-\dfrac{e_{\mathrm p}}{e_{\mathrm p}^{0}},
1-\dfrac{e_{\theta}}{e_{\theta}^{0}}
\right)
\right],
& \text{otherwise}.
\end{cases}
\label{eq:progress_fr3}
\end{align}
Taking the minimum prevents success in either position or orientation alone from being scored as full pose completion.

\subsubsection{Go2 Bipedal Rearing}
The Go2 task requires simultaneous progress in base height and trunk pitch. Its metric is
\begin{align}
p_{\mathrm{Go2}}
=
\operatorname{clip}_{[0,1]}
\left[
\min\left(
\dfrac{z-\underline z}{\overline z-\underline z},
\phi
\right)
\right],
\label{eq:progress_go2}
\end{align}
where $z$ is the base height, $(\underline z,\overline z)=(0.31,0.65)\,\mathrm{m}$, and $\phi\in[0,1]$ is the normalized pitch progress. Here, $\phi=0$ corresponds to the initial four-legged body orientation and $\phi=1$ to the commanded rear-up orientation. The minimum requires both the desired elevation and pitch to be achieved.

\subsubsection{G1 Sustained Squat}
For the G1, progress is measured by the pelvis descent while preserving an upright torso:
\begin{align}
p_{\mathrm{G1}}
=
\operatorname{clip}_{[0,1]}
\left[
\min\left(
\dfrac{z_0-z}{z_0-z^{*}},
u
\right)
\right],
\label{eq:progress_g1}
\end{align}
where $z_0=0.79\,\mathrm{m}$ and $z^{*}=0.62\,\mathrm{m}$ are the initial and desired pelvis heights.
The uprightness term $u\in[0,1]$ is computed from the vertical component of the pelvis up-axis, with $u=1$ for an upright pelvis and $u=0$ once the tilt exceeds $60^{\circ}$.
This term prevents falls that reduce the pelvis height from being scored as successful squats.

The normalized progress metric is used for the simulated comparisons. For the real-world FR3 evaluation, we instead report the end-effector response to the target switch and whether each controller reaches and maintains the newly commanded target.
 
\subsection{In-Distribution: Performance under an Aligned Prior}
\label{sec:id}
\begin{table}[tbp]
\centering
\caption{Simulated In-Distribution Task Progress (Mean $\pm$ std)}
\label{tab:id}
\begin{tabular}{lcc}
\toprule
Injection scheme & 2D navigation & 2D reaching \\
\midrule
Warm-start                                        & $0.988 \pm 0.000$ & $0.991 \pm 0.001$ \\
Mixture                                   & $0.992 \pm 0.001$ & $0.992 \pm 0.001$ \\
Mixture-Elite                             & $0.994 \pm 0.001$ & $0.992 \pm 0.002$ \\
Blend~\cite{huang2024diffusion}            & $0.990 \pm 0.000$ & $0.992 \pm 0.001$ \\
GPC-CEM~\cite{brudermuller2026generative}  & $0.989 \pm 0.001$ & $0.988 \pm 0.002$ \\
Residual-MPPI~\cite{wang2025residualmppi}  & $0.985 \pm 0.000$ & $0.990 \pm 0.001$ \\
ProxPI                                            & $0.989 \pm 0.002$ & $0.985 \pm 0.006$ \\
\bottomrule
\end{tabular}
\end{table}
When the prior is aligned with the deployment task, prior-guided sampling can improve the finite-sample efficiency of MPPI by concentrating rollouts in task-relevant regions, as established in prior work~\cite{power2024learning,kurtz2025generative,brudermuller2026generative}.
Here, we examine whether ProxPI preserves the performance of such an accurate prior before evaluating its robustness to prior mismatch, and whether the site at which the prior is injected matters at all in this regime.
As shown in Table~\ref{tab:id}, all seven injection schemes reach near-optimal task progress on the two ID tasks, and the whole set spans only $0.009$ on navigation and $0.007$ on reaching. At this operating budget the injection site therefore does not change ID performance.
Although ProxPI centers its sampling on the nominal rather than on the prior, the proximal term draws that nominal toward the accurate prior, so its rollouts concentrate in the same region as those of the prior-centered schemes.
The schemes match under an aligned prior and diverge only when the prior is mismatched, which we examine next.

\subsection{Out-of-Distribution: Robustness to Prior Mismatch}
\label{sec:ood}

{\erev
\begin{table*}[tbp]
\centering
\caption{Simulated Out-of-Distribution Task Progress, All Prior-Injection
Schemes (Mean $\pm$ std)}
\label{tab:ood_all}
\setlength{\tabcolsep}{2pt}
\begin{tabular}{lcccccccc}
\toprule
Out-of-distribution Task & Vanilla MPPI & Warm-start & Mixture & Mixture-Elite
 & Blend & GPC-CEM
 & Residual-MPPI & \textbf{ProxPI} \\
\midrule
2D navigation      & $\mathbf{0.997{\pm}0.000}$ & $0.000{\pm}0.000$ & $0.973{\pm}0.139$
 & $0.974{\pm}0.139$ & $0.000{\pm}0.000$ & $\mathbf{0.989{\pm}0.003}$ & {$0.000{\pm}0.000$} & $\mathbf{0.989{\pm}0.002}$ \\
2D reaching        & $\mathbf{0.984{\pm}0.008}$ & $0.000{\pm}0.000$ & $\mathbf{0.979{\pm}0.022}$
 & $\mathbf{0.982{\pm}0.017}$ & $0.000{\pm}0.000$ & $\mathbf{0.979{\pm}0.020}$ & {$0.000{\pm}0.000$} & $\mathbf{0.983{\pm}0.007}$ \\
FR3 obstacle & $\mathbf{0.954{\pm}0.004}$ & $0.000{\pm}0.000$ & $0.931{\pm}0.013$
 & $0.932{\pm}0.012$ & $0.000{\pm}0.000$ & $0.396{\pm}0.413$ & {---} & $\mathbf{0.953{\pm}0.003}$ \\
Go2 rear-up        & $0.759{\pm}0.216$ & $0.000{\pm}0.001$ & $\mathbf{0.894{\pm}0.077}$
 & $0.871{\pm}0.134$ & $0.020{\pm}0.007$ & $0.383{\pm}0.345$ & {---} & $\mathbf{0.895{\pm}0.093}$ \\
G1 squat           & $\mathbf{0.950{\pm}0.033}$ & $0.000{\pm}0.000$ & $0.920{\pm}0.166$
 & $\mathbf{0.948{\pm}0.036}$ & $0.000{\pm}0.000$ & $0.658{\pm}0.433$ & {---} & $\mathbf{0.952{\pm}0.014}$ \\
\bottomrule
\end{tabular}
\par\vspace{2pt}\footnotesize\raggedright
\textbf{Bold} marks the best scheme of a row together with every scheme whose
mean is within $0.01$ of it. References are Blend~\cite{huang2024diffusion},
GPC-CEM~\cite{brudermuller2026generative}, and Residual-MPPI~\cite{wang2025residualmppi}.
Residual-MPPI requires the prior's action likelihood, which among our priors is
available only for the SAC policies of the two low-dimensional tasks, so --- marks
the platforms where it does not apply.\par
\end{table*}
}

When the deployment task differs from the prior's training distribution, the learned prior can become incompatible with the deployment objective or its constraints. We evaluate robustness to such mismatch across five simulated tasks spanning environment and behavior mismatch, and on a real FR3.
The simulated task behaviors} and task-progress scores are reported in Fig.~\ref{fig:ood} and Table~\ref{tab:ood_all}, with the MPPI parameters listed in Table~\ref{tab:config}.

\subsubsection{Simulation}

We first report the learned-prior OOD condition on the two low-DoF platforms.
We then keep these tasks fixed and interpolate the online prior between a high-budget
reference plan and the learned prior. Finally, we evaluate the same OOD contrast on
the three higher-DoF robots.

\paragraph{Low-DoF platforms}
On both the mobile-navigation and planar-reaching tasks, policy-centered warm-start remains concentrated around the prior's direct route and consequently collides with the obstacle, yielding zero progress. By contrast, ProxPI allows the nominal sequence to move away from the mismatched prior and produces collision-free trajectories to the target. It achieves progress scores of $0.989$ for navigation and $0.983$ for planar reaching, matching the corresponding performance of vanilla MPPI, as shown in Fig.~\ref{fig:ood} and Table~\ref{tab:ood_all}.
Blend collides on every seed of both tasks, exactly as warm-start does. It
displaces every rollout toward the prior, so none is sampled around the
nominal.
The mixture schemes and GPC-CEM instead retain rollouts that are not displaced and complete both tasks. 
The larger spread of the mixture schemes on navigation is due to a single collision among $50$ seeds; the remaining $49$ average $0.993$.

Residual-MPPI collides on every seed of both tasks, as warm-start does. It likewise
initializes its proposal from the prior while encouraging agreement with the prior
through the likelihood term. In these tasks, however, the unseen obstacle invalidates
the motion commanded by the prior, and the method does not recover a collision-free
trajectory. This setting creates a direct conflict between the prior behavior that
Residual-MPPI seeks to preserve and the newly imposed obstacle-avoidance requirement.

\paragraph{Controlled Prior Interpolation Under Fixed Tasks}

\begin{figure}[tbp]
\centering
\includegraphics[width=0.9\columnwidth]{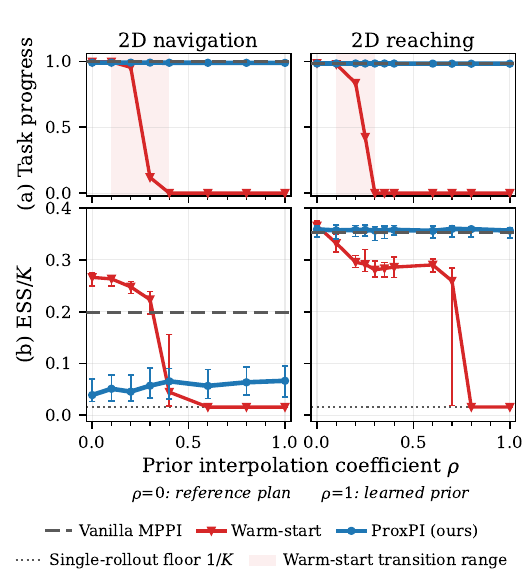}
\caption{Controlled prior interpolation under fixed OOD tasks. (a) Mean task
progress over $N=50$ runs; shading marks the warm-start transition from $50/50$ to
$0/50$ successes. (b) Empirical $\widehat{\mathrm{ESS}}/K$; points and error bars
show the median and interquartile range across per-seed medians over replans. The
dotted line marks the single-rollout floor $1/K$.}
\label{fig:interp}
\end{figure}

The preceding low-DoF results show only the learned-prior OOD endpoint, not how the performance gap between warm-start and ProxPI evolves as the prior is interpolated from a reference plan to the mismatched learned prior.
We therefore hold the OOD navigation and reaching tasks fixed to avoid confounding
prior variation with changes in task difficulty, and construct
\begin{equation}
    U_{\mathrm p}(\rho) = (1-\rho)U_{\mathrm{ref}} + \rho U_{\mathrm{learned}},
\label{eq:interp}
\end{equation}
where $U_{\mathrm{ref}}$ is generated online from the current state by a higher-budget
MPPI solve with $K=128$, and $U_{\mathrm{learned}}$ is the obstacle-unaware learned prior used in the
preceding OOD evaluation. All evaluated controllers use the original $K=64$ operating
point and otherwise retain the same task and controller settings. 
By construction, $\rho=0$ yields the reference plan, whereas $\rho=1$ recovers the
learned-prior OOD condition above.
Because both endpoint sequences are recomputed from the current closed-loop state,
$\rho$ parameterizes the prior-construction rule rather than a calibrated distance to
the task optimum. The higher-budget reference is used only to construct this prior
family and is not treated as a competing controller.
Vanilla MPPI does not use the prior and is therefore unaffected by \(\rho\).

Fig.~\ref{fig:interp}(a) shows that, as $\rho$ increases, warm-start transitions from
successful completion to complete failure over the shaded ranges on both tasks,
whereas ProxPI remains nearly unchanged and succeeds on all $50$ seeds at every tested
coefficient. Within the shaded ranges, the intermediate warm-start scores reflect
mixtures of successful and failed seeds rather than partial completions.
Fig.~\ref{fig:interp}(b) then examines whether empirical ESS tracks this performance
variation. On reaching, warm-start fails on all $50$ seeds from $\rho=0.3$, while its
normalized ESS remains between $0.259$ and $0.290$ through $\rho=0.7$ and reaches the
single-rollout floor only from $\rho=0.8$; navigation shows the same ordering. Task
failure therefore need not coincide with complete weight collapse. ProxPI succeeds
across the interpolation despite substantially different ESS levels between the two
tasks, so the absolute ESS level is not a task-independent success threshold in these
experiments either.

To quantify the reset effect directly, at each replan $t$ we define
$\Delta J_{\mathrm{reset},t}(\rho)=S_t(U_{\mathrm{p},t}(\rho))-S_t(U_t^-)$, where
$U_{\mathrm{p},t}(\rho)$ is the interpolated prior of \eqref{eq:interp} constructed at
the current replan, and $U_t^-$ is the nominal carried from the previous replan using
the controller's receding-horizon shift. Both are evaluated from the same current
state under the common task cost, excluding the proximal term. Thus,
$\Delta J_{\mathrm{reset},t}(\rho)>0$ indicates that the prior has a higher task cost
than the carried nominal. At $\rho=0.3$, the fraction of replans satisfying this
condition, computed per seed and then averaged across seeds, is $94\%$ for navigation
and $71\%$ for reaching. This provides a closed-loop counterpart to the second
mechanism of Sec.~\ref{sec:finite_sample}, showing that re-centering can replace a
lower-task-cost carried nominal even before the ESS in Fig.~\ref{fig:interp}(b)
collapses.

\paragraph{Higher-DoF platforms}
The three higher-DoF platforms exhibit the same contrast, as shown in Fig.~\ref{fig:ood} and Table~\ref{tab:ood_all}. Policy-centered warm-start obtains zero progress on every task: the FR3
collides with the obstacle, while the Go2 and G1 remain close to the motions prescribed by their locomotion priors and fail to initiate the commanded rearing and squat, respectively. Blend fails with it, reaching at most $0.020$ on any of the three. In contrast, ProxPI allows the MPPI nominal to move away from the mismatched prior and achieves progress scores of $0.953$, $0.895$, and $0.952$ on the FR3, Go2, and G1, respectively.
ProxPI matches vanilla MPPI on the FR3 and the G1 and exceeds it by $0.136$ on the Go2. The gain appears only there because vanilla is already near its ceiling on the other two, at $0.954$ and $0.950$, while on the Go2 it reaches only $0.759$.
The mixture schemes, which retain a prior-free half, trail ProxPI by $0.001$ to $0.032$ across the three platforms.

GPC-CEM is the exception, completing both low-DoF tasks but reaching only
$0.396$, $0.383$ and $0.658$ on the FR3, Go2 and G1. This does not contradict the
results reported for the method, which were obtained under task variations that
leave the prior usable, such as a changed object shape or an added obstacle cost,
rather than under a prior whose commanded motion cannot complete the task at all.
Under the mismatch used here, the shortfall does not follow from the sample
split, since GPC-CEM commits the same half of its budget to the prior as the
mixture schemes, nor from the elite selection, which Mixture-Elite also applies
while remaining close to ProxPI. It differs from both schemes in its
cross-entropy update, which executes the single lowest-cost candidate rather than
a weighted average over the rollouts and contracts the sampling covariance onto
the elite set.

Across all five simulated platforms, the two schemes that leave no rollout at the nominal, warm-start and Blend, collapse to near-zero progress, whereas ProxPI recovers vanilla MPPI-level performance under both environment and behavior mismatch.
This consistency across robot morphologies, action spaces, and task objectives indicates that the failure stems from displacing every rollout away from the nominal rather than from a particular platform, task, or scheme.
Having established this behavior under controlled simulation, we next evaluate the same contrast on real hardware.

\begin{figure}[tbp]
\centering
\includegraphics[width=0.95\columnwidth]{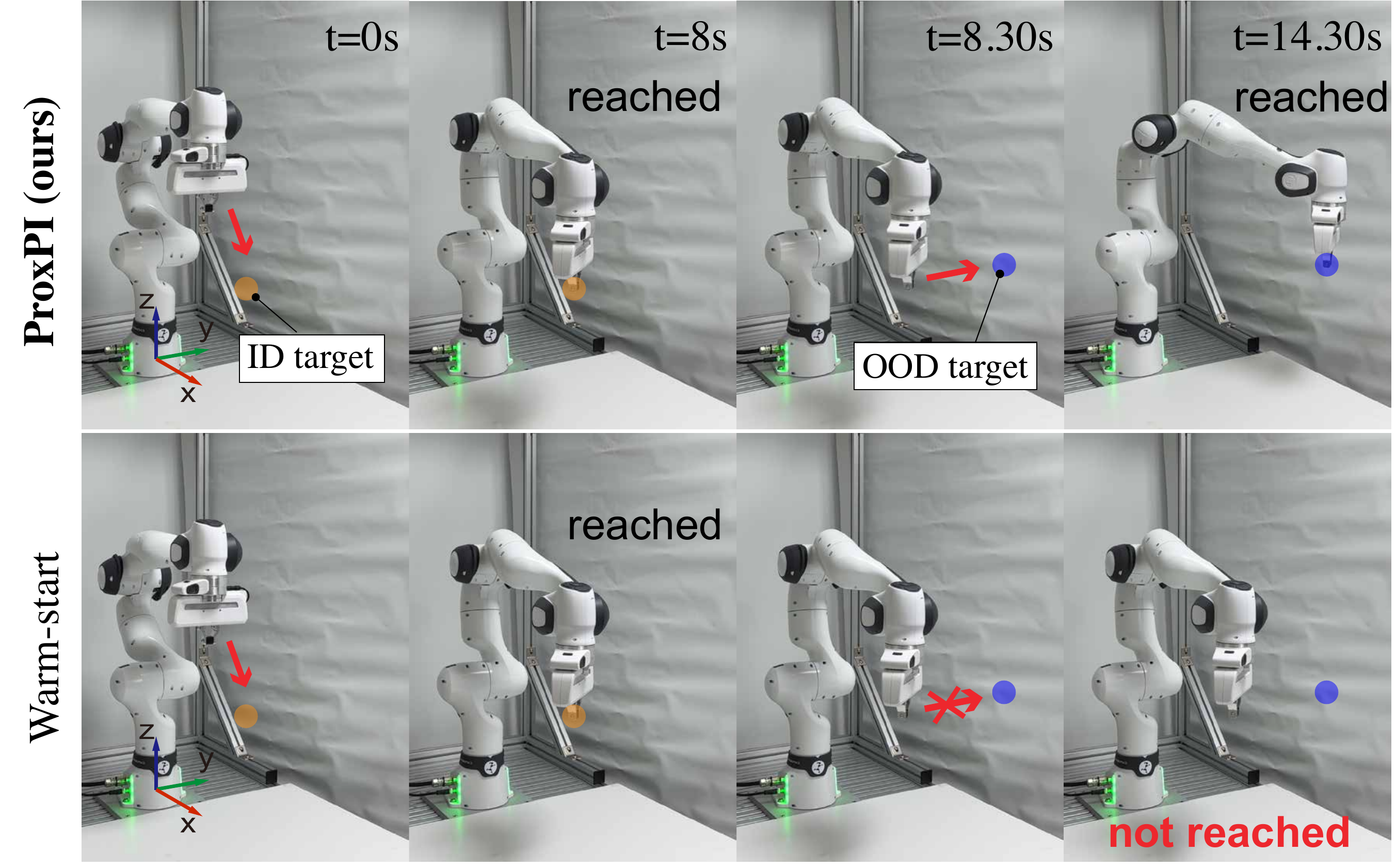}
\caption{Real-world FR3 reaching with a single-target behavioral-cloning prior. Both methods first reach the in-distribution target at $t{=}8$\,s.
After the commanded target switches to an out-of-distribution location at $t{=}8.3$\,s, ProxPI (top) adapts online and reaches the new target by $t{=}14.3$\,s, whereas warm-start (bottom) remains near the training target and fails to reach the new one.}
\label{fig:ood_real_world_fr3}
\end{figure}

\subsubsection{Real-World Experiment}
 
We next evaluate whether the same response to prior mismatch persists on a Franka FR3 arm.
Both controllers first converge to the ID target represented by the behavioral-cloning prior. When the commanded target is subsequently switched outside the prior's training distribution, their behaviors diverge markedly, as shown in Fig.~\ref{fig:ood_real_world_fr3}. Policy-centered warm-start remains near the trained target and fails to respond to the new command, because its sampling center continues to be reset by the prior. In contrast, ProxPI moves away from the trained behavior and reaches the new target by allowing the task cost to override the mismatched prior. This hardware result demonstrates that ProxPI retains the ability to adapt online when the deployment command lies outside the learned prior's training distribution.

\begin{figure}[tbp]
\centering
\includegraphics[width=0.41\textwidth]{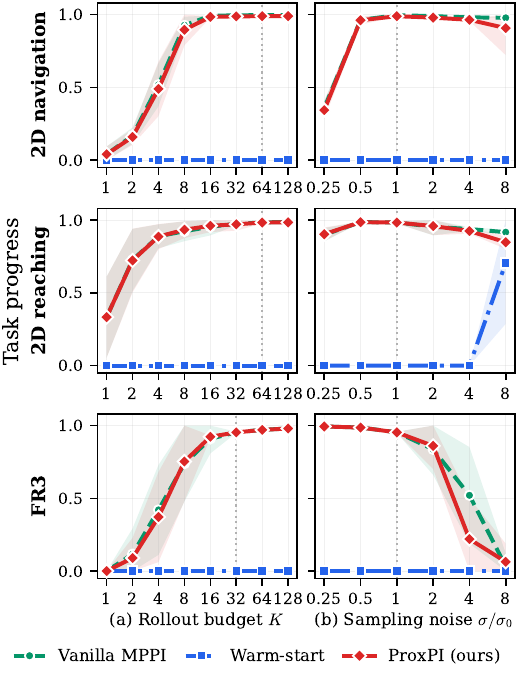}
\caption{Sensitivity to the rollout budget $K$ and sampling-noise scale under OOD prior mismatch. (a) Rollout budget $K$ and (b) normalized sampling-noise scale $\sigma/\sigma_0$, with all other parameters held fixed at the task-specific operating values in Table~\ref{tab:config}.
Curves show the mean, and shaded regions denote $\pm1$ standard deviation.}
\label{fig:ood_sweeps}
\end{figure}

We next analyze parameter sensitivity and finite-sample weight concentration
under OOD prior mismatch. Throughout these analyses, all parameters other than the
one being varied are held at their operating values in Table~\ref{tab:config}; when
the sampling-noise scale is varied, $\sigma$ is normalized by its task-specific
operating value $\sigma_0$.

\subsection{Parameter Sensitivity}
The sweeps in Fig.~\ref{fig:ood_sweeps} show that increasing the number or spread of samples does not consistently resolve warm-start failure under OOD prior mismatch.
In Fig.~\ref{fig:ood_sweeps}(a), increasing $K$ improves vanilla MPPI and ProxPI, whereas warm-start remains near zero on all three tasks even at the largest budget, indicating that additional samples around the mismatched prior do not resolve the failure.
In Fig.~\ref{fig:ood_sweeps}(b), increasing $\sigma/\sigma_0$ allows warm-start to escape the prior only at extreme noise levels on the 2D reaching task, but not on 2D navigation or FR3, while excessive noise degrades vanilla MPPI and ProxPI on FR3.
Thus, broadening the sampling distribution can overcome prior mismatch in some lower-dimensional settings, but does not provide a reliable solution across tasks.

\begin{figure}[tbp]
\centering
\includegraphics[width=0.8\columnwidth]{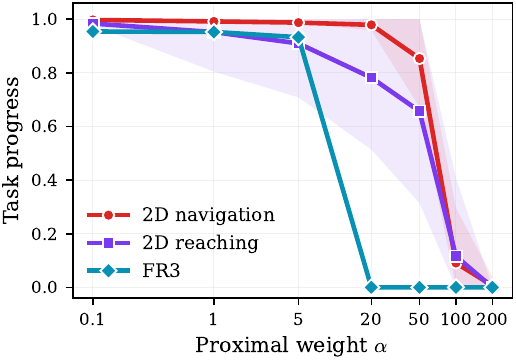} 
\caption{Sensitivity of ProxPI to the proximal weight $\alpha$ under OOD prior mismatch, with all other parameters fixed at the operating values in Table~\ref{tab:config}. Curves show the mean, and shaded regions denote $\pm1$ standard deviation.}
\label{fig:ood_alpha}
\end{figure}

Fig.~\ref{fig:ood_alpha} shows how the proximal weight $\alpha$ controls the influence of the learned prior.
ProxPI maintains high task progress for small to moderate $\alpha$, but degrades and eventually collapses as the prior is weighted too strongly.
This degradation occurs at smaller $\alpha$ on FR3 than on the two low-dimensional tasks, suggesting that the allowable strength of prior regularization may be more restrictive for the higher-dimensional manipulation setting.

\subsection{Rollout Budget across Injection Schemes}
\begin{figure}[tbp]
\centering
\includegraphics[width=\columnwidth]{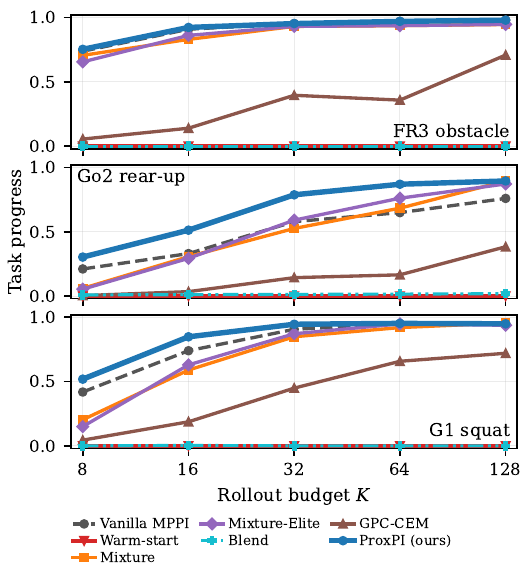}
\caption{Rollout-budget comparison of vanilla MPPI and all six injection
schemes on the three higher-DoF platforms under OOD prior mismatch.
Each scheme runs at its own best $\lambda$. Curves show the mean; Table~\ref{tab:ood_all} reports the spread over seeds at the operating points. The warm-start and Blend curves coincide near zero on every panel.}
\label{fig:budget}
\end{figure}

Fig.~\ref{fig:budget} extends the budget sweep to every injection scheme on the
three higher-DoF platforms.
Warm-start and Blend stay at zero progress for every budget tested, so their
failure cannot be attributed to an insufficient sample budget.
GPC-CEM improves steadily with $K$ but remains far below the other schemes
throughout.
The mixture schemes trail the others most at small budgets, where committing half of the
rollouts to a mismatched prior is most costly, and catch up as $K$ grows: on the
Go2 the gap between ProxPI and Mixture falls from $0.243$ at $K=8$ to 
around zero at $K=128$.
ProxPI is never more than $0.003$ below vanilla MPPI at any budget and exceeds it
by $0.136$ to $0.221$ on the Go2 for $K\ge16$. On the G1 it leads at the smallest
budgets and the two converge by $K=64$.
Taken together, these results show that increasing $K$ does not rescue the schemes
that displace every rollout away from the nominal, while the performance
differences among the remaining schemes are most pronounced in the small-budget
regime that motivates prior injection.

\subsection{Weight Concentration under a Common Task Cost}
\label{sec:ess}
\begin{figure}[tbp]
\centering
\includegraphics[width=\columnwidth]{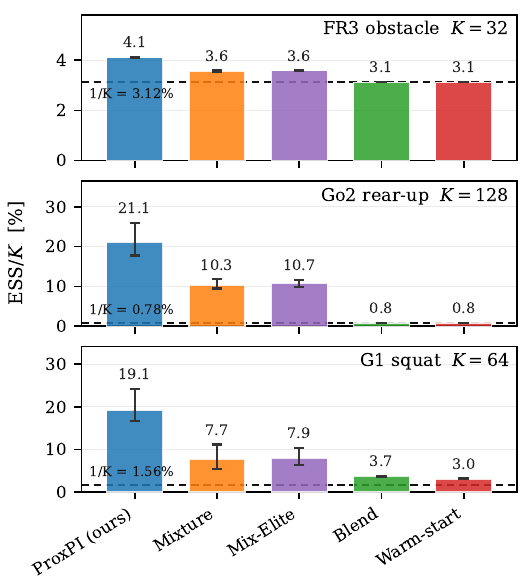}
\caption{Empirical $\widehat{\mathrm{ESS}}/K$ at the OOD operating points of
the three higher-DoF platforms. Bars show the median across seeds of the
per-replan medians, error bars the interquartile range, and dashed lines the
single-rollout floor $1/K$. Weights are computed from the task cost $S$ alone at a
common temperature within each task. GPC-CEM is omitted because its rank-based
elite selection forms no softmax weights.}
\label{fig:ess}
\end{figure}

Sec.~\ref{sec:finite_sample} identifies proposal--target displacement as a source
of finite-sample weight concentration. We therefore examine how the
softmax-weighted prior-injection schemes concentrate their rollout weights at the
OOD operating points of the three higher-DoF platforms. To isolate the effect of
sampling placement from scheme-specific objective terms, we evaluate all schemes
using the task cost $S$ alone, excluding the proximal term, with a common
temperature within each task. Fig.~\ref{fig:ess} reports the resulting
$\widehat{\mathrm{ESS}}/K$ of \eqref{eq:ess}. Since $\widehat{\mathrm{ESS}}=1$ when
a single rollout carries all the weight, the normalized measure has a
single-rollout floor of $1/K$, marked in each panel.

On the Go2, warm-start and Blend are effectively at the single-rollout floor, with
$\widehat{\mathrm{ESS}}/K=0.8\%$ against $1/K=0.78\%$. The FR3 shows a similar
pattern for these two schemes, with both at $3.1\%$, effectively at the $3.12\%$
floor. On the G1, they remain the two most concentrated schemes, at $3.0\%$ and
$3.7\%$ against a floor of $1.56\%$.
The mixture schemes occupy an intermediate regime, retaining $10.3\%$ and $10.7\%$
of the rollout budget on the Go2 against ProxPI's $21.1\%$, and $7.7\%$ and $7.9\%$
on the G1 against $19.1\%$.
ProxPI retains the largest effective budget among the injection schemes on all
three platforms.
On the FR3, however, all schemes remain close to the floor, so ESS provides little
separation among them at this operating point.
Where the diagnostic does separate the schemes, the observed pattern is consistent
with their sampling placement: ProxPI draws all rollouts from the nominal-centered
proposal, the mixture schemes draw half of their candidate pool from it, whereas
warm-start and Blend draw none from the nominal-centered proposal.

\section{CONCLUSIONS}

In this paper, we proposed Proximal Prior Injection (ProxPI), which preserves the evolving MPPI nominal as the sampling center and incorporates a learned prior through a proximal cost, matching conventional prior-centered sampling schemes under aligned priors while providing greater robustness under prior mismatch.
We showed both theoretically and empirically why this distinction matters: nominal-centered updates retain corrections across replans, whereas policy-centered re-centering discards them; under a local Gaussian approximation, we established this repeated-update mechanism and distinguished it from finite-sample weight concentration.

Under aligned priors, ProxPI matched the performance of the other prior-injection schemes on the simulated ID tasks. 
When the prior became mismatched, ProxPI recovered vanilla MPPI-level performance across five simulated platforms and successfully adapted to the shifted target on a real FR3.
Of the schemes evaluated on all five platforms, ProxPI was the only one within $0.01$ of the best result on every platform.
The ablations further showed that increasing the rollout budget or sampling variance does not reliably recover warm-start, and that excessive proximal weighting can again restrict adaptation.
In the weight analysis, warm-start and Blend yielded the smallest effective sample sizes on the higher-DoF platforms under a common task-cost weighting, effectively reaching the single-rollout floor on the FR3 and Go2, while ProxPI retained the largest effective budget among the injection schemes.

In future work, we plan to develop adaptive prior weighting that adjusts the proximal influence according to the task and degree of prior mismatch, and to extend the theoretical analysis beyond the local Gaussian setting to more general sampling distributions.




\appendices
\section{Local Gibbs Target and Fixed Point}
\label{app:local_target}

Substituting the local model~\eqref{eq:local_cost} into~\eqref{eq:mppi_optdist}
with $q(V)=\mathcal{N}(\bar U,\Sigma)$,
\begin{align}
\log p_\alpha(V)
&=
-\frac{1}{2\lambda}(V-U^*)^\top H(V-U^*)
-\frac{\alpha}{\lambda}\lVert V-U_{\mathrm p}\rVert^2
\nonumber\\
&\quad
-\frac{1}{2}(V-\bar U)^\top\Sigma^{-1}(V-\bar U)
+\mathrm{const},
\end{align}
where the constant collects all terms independent of $V$.
Collecting the quadratic and linear terms in $V$ gives the precision
$\Sigma^{-1}+(H+2\alpha I)/\lambda$ and the linear coefficient
$\Sigma^{-1}\bar U+\lambda^{-1}HU^*+2\alpha\lambda^{-1}U_{\mathrm p}$, which are
\eqref{eq:local_cov} and~\eqref{eq:local_mean}. For nominal-centered sampling,
setting
$m_\alpha(U^\dagger)=U^\dagger$ and cancelling $\Sigma^{-1}U^\dagger$ from both
sides leaves $(H+2\alpha I)U^\dagger=HU^*+2\alpha U_{\mathrm p}$, which is
\eqref{eq:prox_fixed_point}.

\section{Proof of Proposition~\ref{prop:contraction}}
\label{app:contraction}

From~\eqref{eq:local_mean} and~\eqref{eq:prox_fixed_point},
\begin{align}
m_\alpha(U)-U^\dagger
&=
C_\alpha\Sigma^{-1}(U-U^\dagger)
\nonumber\\
&=
A_\alpha(U-U^\dagger),
\end{align}
which establishes~\eqref{eq:prox_contraction}.

Moreover, $H+2\alpha I\succ0$ and~\eqref{eq:local_cov} imply
\begin{equation}
C_\alpha^{-1}
=
\Sigma^{-1}
+
\frac{H+2\alpha I}{\lambda}
\succ
\Sigma^{-1}.
\end{equation}
Therefore
\begin{equation}
C_\alpha^{1/2}
\Sigma^{-1}
C_\alpha^{1/2}
\prec I.
\end{equation}
The matrix $A_\alpha=C_\alpha\Sigma^{-1}$ is similar to
$C_\alpha^{1/2}\Sigma^{-1}C_\alpha^{1/2}$, which is symmetric positive
definite. Hence all eigenvalues of $A_\alpha$ lie strictly between zero
and one, and therefore $\rho(A_\alpha)<1$.
\hfill$\square$

\section{Gaussian Chi-Squared Divergence}
\label{app:gaussian_chi2}

Consider two Gaussian distributions
\begin{equation}
p(V)=\mathcal{N}(m,C),
\qquad
q(V)=\mathcal{N}(\mu,\Sigma),
\end{equation}
and define $g=m-\mu$. By definition,
\begin{equation}
\chi^2(p\|q)+1
=
\int \frac{p(V)^2}{q(V)}\,dV.
\label{eq:app_chi_def}
\end{equation}
With the change of variables $x=V-\mu$,
\begin{align}
\frac{p(V)^2}{q(V)}
&=
(2\pi)^{-d/2}
\frac{|\Sigma|^{1/2}}{|C|}
\nonumber\\
&\quad\times
\exp\!\left[
-(x-g)^\top C^{-1}(x-g)
+\frac{1}{2}x^\top\Sigma^{-1}x
\right].
\label{eq:app_ratio}
\end{align}
Define
\begin{equation}
D
=
2C^{-1}-\Sigma^{-1},
\qquad
h
=
2C^{-1}g.
\end{equation}
The exponent in~\eqref{eq:app_ratio} becomes
\begin{equation}
-\frac{1}{2}x^\top D x
+
h^\top x
-
g^\top C^{-1}g.
\end{equation}
Completing the square gives
\begin{align}
&-\frac{1}{2}
(x-D^{-1}h)^\top
D
(x-D^{-1}h)
\nonumber\\
&\qquad
+
\frac{1}{2}h^\top D^{-1}h
-
g^\top C^{-1}g.
\end{align}
When $D\succ0$, integrating the Gaussian term yields
\begin{align}
\chi^2(p\|q)+1
&=
\frac{|\Sigma|^{1/2}}
{|C|\,|D|^{1/2}}
\nonumber\\
&\quad\times
\exp\!\left[
\frac{1}{2}h^\top D^{-1}h
-
g^\top C^{-1}g
\right].
\label{eq:app_chi_intermediate}
\end{align}

Using
\begin{equation}
D
=
C^{-1}(2\Sigma-C)\Sigma^{-1},
\end{equation}
its determinant satisfies
\begin{equation}
|D|
=
\frac{|2\Sigma-C|}
{|C|\,|\Sigma|}.
\end{equation}
Furthermore,
\begin{equation}
D^{-1}
=
\Sigma(2\Sigma-C)^{-1}C,
\end{equation}
which gives
\begin{equation}
\frac{1}{2}h^\top D^{-1}h
-
g^\top C^{-1}g
=
g^\top(2\Sigma-C)^{-1}g.
\end{equation}
Substituting these identities into~\eqref{eq:app_chi_intermediate} gives
\begin{equation}
\chi^2(p\|q)+1
=
\frac{|\Sigma|}
{\sqrt{|C|\,|2\Sigma-C|}}
\exp\!\left[
g^\top(2\Sigma-C)^{-1}g
\right].
\label{eq:app_general_chi2}
\end{equation}

Setting
$C=C_\alpha$,
$m=m_\alpha(\bar U)$,
$\mu=\bar U$, and
$g=g_\alpha$
recovers~\eqref{eq:general_chi2}.
Under the local model,
$C_\alpha\preceq\Sigma$, which ensures the required Gaussian integral is
finite.
\hfill$\square$


\section{Experimental Details}
\label{app:exp_details}
The per-joint sampling standard deviations are $0.05$ for FR3 joints~1--4 and $0.04$
for joints~5--7. On the G1 they follow the same grouping as the whole-body controller,
with $0.030$ for the hip-pitch, knee, and ankle-pitch joints, $0.025$ for the remaining
leg joints, $0.015$ for the waist, and $0.010$ for the arms, since a uniform value
destabilizes the arms. The Go2 uses a uniform $0.10$.
For the Go2 and G1, holding the policy outputs constant over the MPPI horizon avoids
repeated per-node policy evaluations and reduces the associated planning overhead.

The ACT prior used for the real-robot FR3 evaluation follows the LeRobot reference
implementation~\cite{cadene2024lerobot} of ACT~\cite{zhao2023learning}, with the vision
inputs removed. It observes only the FR3 joint positions and velocities and predicts a
$40$-step joint-position chunk that is used directly as $U_{\mathrm p}$. Training
demonstrations are generated in MuJoCo for a single
end-effector target at $(0.5,0,0.336)\,\mathrm{m}$ with fixed orientation, and the
prior takes no target input, so it drives the end-effector toward the training target
irrespective of the commanded one. During
evaluation, after the robot converges to this target, the commanded position is
switched to $(0.4,0.35,0.336)\,\mathrm{m}$ while the commanded orientation is left
unchanged.

All simulations are run on a 20-core Intel Core Ultra~7 265KF CPU.
For the real-world experiments, the Franka Control Interface runs on an Intel NUC 13 equipped with a Core i7 processor and 32 GB of RAM.

\bibliographystyle{IEEEtran}
\bibliography{MyReference} 

@article{williams2017model,
  title={Model predictive path integral control: From theory to parallel computation},
  author={Williams, Grady and Aldrich, Andrew and Theodorou, Evangelos A},
  journal={Journal of Guidance, Control, and Dynamics},
  volume={40},
  number={2},
  pages={344--357},
  year={2017},
  publisher={American Institute of Aeronautics and Astronautics}
}

@article{williams2018information,
  title={Information-theoretic model predictive control: Theory and applications to autonomous driving},
  author={Williams, Grady and Drews, Paul and Goldfain, Brian and Rehg, James M and Theodorou, Evangelos A},
  journal={IEEE Transactions on Robotics},
  volume={34},
  number={6},
  pages={1603--1622},
  year={2018},
  publisher={IEEE}
}

@article{howell2022predictive,
  title={Predictive sampling: Real-time behaviour synthesis with mujoco},
  author={Howell, Taylor and Gileadi, Nimrod and Tunyasuvunakool, Saran and Zakka, Kevin and Erez, Tom and Tassa, Yuval},
  journal={arXiv preprint arXiv:2212.00541},
  year={2022}
}

@article{kim2025single,
  title={Single-Instance Sampling for Computationally Efficient and Accurate Real-Time Task Space MPPI Control},
  author={Kim, Dongwhan and Im, Euncheol and Kim, Yujin and Lim, Myotaeg and Lee, Yisoo},
  journal={IEEE Transactions on Robotics},
  volume={41},
  pages={6327--6344},
  year={2025},
  publisher={IEEE}
}

@article{power2024learning,
  title={Learning a generalizable trajectory sampling distribution for model predictive control},
  author={Power, Thomas and Berenson, Dmitry},
  journal={IEEE Transactions on Robotics},
  volume={40},
  pages={2111--2127},
  year={2024},
  publisher={IEEE}
}

@inproceedings{sacks2023learning,
  title={Learning sampling distributions for model predictive control},
  author={Sacks, Jacob and Boots, Byron},
  booktitle={Conference on Robot Learning},
  pages={1733--1742},
  year={2023},
  organization={PMLR}
}

@inproceedings{huang2024diffusion,
  title={Diffusion-MPPI: Diffusion Informed Model Predictive Path Integral Method},
  author={Huang, Yi and Liu, Houde},
  booktitle={International Conference on Neural Information Processing},
  pages={284--298},
  year={2024},
  organization={Springer}
}

@article{kotecha2025real,
  title={Real-Time Gait Adaptation for Quadrupeds using Model Predictive Control and Reinforcement Learning},
  author={Kotecha, Prakrut and Kolathaya, Shishir and others},
  journal={arXiv preprint arXiv:2510.20706},
  year={2025}
}

@article{kurtz2025generative,
  title={Generative predictive control: Flow matching policies for dynamic and difficult-to-demonstrate tasks},
  author={Kurtz, Vince and Burdick, Joel W},
  journal={arXiv preprint arXiv:2502.13406},
  year={2025}
}

@article{brudermuller2026generative,
  title={Generative Models from and for Sampling-Based MPC: A Bootstrapped Approach for Adaptive Contact-Rich Manipulation},
  author={Bruderm{\"u}ller, Lara and Hung, Brandon and Zhu, Xinghao and Wang, Jiuguang and Hawes, Nick and Culbertson, Preston and Le Cleac'h, Simon},
  journal={IEEE Robotics and Automation Letters},
  year={2026},
  publisher={IEEE}
}

@article{cheng2025rambo,
  title={{RAMBO}: {RL}-Augmented Model-Based Whole-Body Control for Loco-Manipulation},
  author={Cheng, Jin and Kang, Dongho and Fadini, Gabriele and Shi, Guanya and Coros, Stelian},
  journal={IEEE Robotics and Automation Letters},
  year={2025},
  publisher={IEEE}
}

@article{jeon2025residual,
  title={Residual {MPC}: Blending Reinforcement Learning with {GPU}-Parallelized Model Predictive Control},
  author={Jeon, Se Hwan and Lee, Ho Jae and Hong, Seungwoo and Kim, Sangbae},
  journal={arXiv preprint arXiv:2510.12717},
  year={2025}
}

@inproceedings{zhang2016mpcgps,
  title={Learning deep control policies for autonomous aerial vehicles with {MPC}-guided policy search},
  author={Zhang, Tianhao and Kahn, Gregory and Levine, Sergey and Abbeel, Pieter},
  booktitle={2016 IEEE International Conference on Robotics and Automation (ICRA)},
  year={2016},
  organization={IEEE}
}

@article{kong1992note,
  title={A note on importance sampling using standardized weights},
  author={Kong, Augustine},
  journal={University of Chicago, Dept. of Statistics, Tech. Rep},
  volume={348},
  pages={14},
  year={1992}
}

@article{agapiou2017importance,
  title={Importance sampling: Intrinsic dimension and computational cost},
  author={Agapiou, Sergios and Papaspiliopoulos, Omiros and Sanz-Alonso, Daniel and Stuart, Andrew M},
  journal={Statistical Science},
  pages={405--431},
  year={2017},
  publisher={JSTOR}
}

@inproceedings{yi2024covo,
  title={CoVO-MPC: Theoretical analysis of sampling-based MPC and optimal covariance design},
  author={Yi, Zeji and Pan, Chaoyi and He, Guanqi and Qu, Guannan and Shi, Guanya},
  booktitle={6th Annual Learning for Dynamics \& Control Conference},
  pages={1122--1135},
  year={2024},
  organization={PMLR}
}

@inproceedings{rudin2022learning,
  title={Learning to walk in minutes using massively parallel deep reinforcement learning},
  author={Rudin, Nikita and Hoeller, David and Reist, Philipp and Hutter, Marco},
  booktitle={Conference on robot learning},
  pages={91--100},
  year={2022},
  organization={PMLR}
}

@article{hwangbo2019learning,
  title={Learning agile and dynamic motor skills for legged robots},
  author={Hwangbo, Jemin and Lee, Joonho and Dosovitskiy, Alexey and Bellicoso, Dario and Tsounis, Vassilios and Koltun, Vladlen and Hutter, Marco},
  journal={Science robotics},
  volume={4},
  number={26},
  pages={eaau5872},
  year={2019},
  publisher={American Association for the Advancement of Science}
}

@article{kumar2021rma,
  title={Rma: Rapid motor adaptation for legged robots},
  author={Kumar, Ashish and Fu, Zipeng and Pathak, Deepak and Malik, Jitendra},
  journal={arXiv preprint arXiv:2107.04034},
  year={2021}
}

@misc{unitree_rl_lab,
  author       = {{Unitree Robotics}},
  title        = {Unitree {RL} {Lab}: Reinforcement Learning for Unitree Robots},
  howpublished = {\url{https://github.com/unitreerobotics/unitree_rl_lab}},
  note         = {GitHub repository, accessed 2026-07-05}
}

@article{qu2024rldriven,
  author  = {Qu, Yue and Chu, Hongqing and Gao, Shuhua and Guan, Jun
             and Yan, Haoqi and Xiao, Liming and Li, Shengbo Eben
             and Duan, Jingliang},
  title   = {{RL}-Driven {MPPI}: Accelerating Online Control Laws
             Calculation With Offline Policy},
  journal = {IEEE Transactions on Intelligent Vehicles},
  year    = {2024},
  volume  = {9},
  number  = {2},
  pages   = {3605--3616},
  doi     = {10.1109/TIV.2023.3348134}
}

@article{seo2026rgb,
  author  = {Seo, Yunsoo and Choi, Sol and Im, Euncheol
             and Lim, Myo Taeg and Lee, Yisoo},
  title   = {{RGB}: {RL} Guided Whole-Body {MPPI} for Humanoid Control},
  journal = {arXiv preprint arXiv:2606.25123},
  year    = {2026},
  doi     = {10.48550/arXiv.2606.25123},
  eprint  = {2606.25123},
  archiveprefix = {arXiv},
  primaryclass  = {cs.RO}
}

@inproceedings{wang2025residualmppi,
  author    = {Wang, Pengcheng and Li, Chenran and Weaver, Catherine
               and Kawamoto, Kenta and Tomizuka, Masayoshi
               and Tang, Chen and Zhan, Wei},
  title     = {Residual-{MPPI}: Online Policy Customization
               for Continuous Control},
  booktitle = {International Conference on Learning Representations},
  year      = {2025},
  url       = {https://openreview.net/forum?id=gVnJFY8nCM}
}

@misc{cadene2024lerobot,
    author = {Cadene, Remi and Alibert, Simon and Soare, Alexander and Gallouedec, Quentin and Zouitine, Adil and Palma, Steven and Kooijmans, Pepijn and Aractingi, Michel and Shukor, Mustafa and Aubakirova, Dana and Russi, Martino and Capuano, Francesco and Pascal, Caroline and Choghari, Jade and Meftah, Khalil and Ellerbach, Maxime and Moss, Jess and Wolf, Thomas},
    title = {LeRobot: State-of-the-art Machine Learning for Real-World Robotics in Pytorch},
    howpublished = "\url{https://github.com/huggingface/lerobot}",
    year = {2024}
}

@inproceedings{zhao2023learning,
    title     = {Learning Fine-Grained Bimanual Manipulation with Low-Cost Hardware},
    author    = {Zhao, Tony Z. and Kumar, Vikash and Levine, Sergey and Finn, Chelsea},
    booktitle = {Proceedings of Robotics: Science and Systems (RSS)},
    year      = {2023}
}

@inproceedings{yoon2022sampling,
  title={Sampling complexity of path integral methods for trajectory optimization},
  author={Yoon, Hyung-Jin and Tao, Chuyuan and Kim, Hunmin and Hovakimyan, Naira and Voulgaris, Petros},
  booktitle={American Control Conference},
  pages={3482--3487},
  year={2022},
  organization={IEEE}
}

@article{honda2026inference,
  title={Model Predictive Control via Probabilistic Inference: A Tutorial and Survey},
  author={Honda, Kohei},
  journal={Annual Reviews in Control},
  volume={61},
  pages={101052},
  year={2026}
}

@inproceedings{johannink2019residual,
  title={Residual reinforcement learning for robot control},
  author={Johannink, Tobias and Bahl, Shikhar and Nair, Ashvin and Luo, Jianlan and Kumar, Avinash and Loskyll, Matthias and Ojea, Juan Aparicio and Solowjow, Eugen and Levine, Sergey},
  booktitle={IEEE International Conference on Robotics and Automation},
  pages={6023--6029},
  year={2019},
  organization={IEEE}
}

@article{liang2026policy,
  title={Policy-guided model predictive path integral for safe manipulator trajectory planning},
  author={Liang, L. and Wu, C. and Wang, X.},
  journal={Sensors},
  volume={26},
  number={7},
  pages={2074},
  year={2026}
}

@article{trevisan2024biased,
  title={Biased-{MPPI}: Informing sampling-based model predictive control by fusing ancillary controllers},
  author={Trevisan, Elia and Alonso-Mora, Javier},
  journal={IEEE Robotics and Automation Letters},
  volume={9},
  number={6},
  pages={5871--5878},
  year={2024}
}

@article{carius2022constrained,
  title={Constrained stochastic optimal control with learned importance sampling: A path integral approach},
  author={Carius, Jan and Ranftl, Ren{\'e} and Farshidian, Farbod and Hutter, Marco},
  journal={The International Journal of Robotics Research},
  volume={41},
  number={2},
  pages={189--209},
  year={2022}
}

\end{document}